%% file: main.tex
\pdfoutput=1

\documentclass[conference]{IEEEtran}

\usepackage{cite}
\usepackage{amssymb}
\usepackage{amsmath}
\usepackage{mathdots}
\usepackage[linesnumbered,ruled,vlined]{algorithm2e}
\SetInd{0.1em}{0.75em}
\SetVlineSkip{0.1em}
\SetAlFnt{\small}
\usepackage{xfp}
\usepackage{tikz}
\usetikzlibrary{arrows,angles,quotes,calc,shapes}
\usepackage{pgfplots,xparse}
\usepackage[colorlinks,bookmarksopen,bookmarksnumbered,citecolor=blue,urlcolor=blue]{hyperref}
\usepackage[autostyle=false]{csquotes}
\usepackage{booktabs}
\usepackage{subcaption}
\usepackage{mathtools}
\mathtoolsset{showonlyrefs}
\usepackage{listings}
\usepackage{float}
\usepackage{multicol}
\usepackage{stfloats}

\pgfplotsset{compat=newest}
\usetikzlibrary{decorations.markings}
\usetikzlibrary{arrows.meta}
\usetikzlibrary{shapes.misc}
\usetikzlibrary{patterns}
\usepgfplotslibrary{groupplots}
\usepgfplotslibrary{units}

\usetikzlibrary{external}
\usepgfplotslibrary{external}
\newtheorem{definition}{Definition}

\DeclareMathOperator*{\argmin}{arg\,min}

\newcommand{\TODO}[1]{}

\let\originalleft\left
\let\originalright\right
\renewcommand{\left}{\mathopen{}\mathclose\bgroup\originalleft}
\renewcommand{\right}{\aftergroup\egroup\originalright}

\begin{document}
  \title{Batch Informed Trees (BIT*)}

  \author{\IEEEauthorblockN{James Swedeen}
    \IEEEauthorblockA{Department of Electrical and Computer Engineering\\
                      Utah State University\\
                      Logan, Utah 84322\\
                      Email: james.swedeen@usu.edu}
    \and
    \IEEEauthorblockN{Greg Droge}
    \IEEEauthorblockA{Department of Electrical and Computer Engineering\\
                      Utah State University\\
                      Logan, Utah 84322\\
                      Email: greg.droge@usu.edu}
    }

  \maketitle

  \begin{abstract}
    Path planning through complex obstacle spaces is a fundamental requirement of many mobile robot applications.
    Recently a rapid convergence path planning algorithm, Batch Informed Trees (BIT*), was introduced.
    This work serves as a concise write-up and explanation of BIT*.
    This work includes a description of BIT* and how BIT* operates, a graphical demonstration of BIT*, and simulation results where BIT* is compared to Optimal Rapidly-exploring Random Trees (RRT*).
  \end{abstract}

  \section{Introduction}
    The ability to plan paths through complex obstacles is a fundamental requirement of many mobile robot applications and is an NP-complete problem in general \cite{Lavalle2006}.
    The literature has seen an explosive growth in sampling-based motion planning algorithms \cite{Gammell2014, Hussain2015, Nasir2013, Noreen2016, Yang2014} that provide asymptotic guarantees for this NP-complete problem.
    These algorithms operate on the principles of dynamic programming, breaking the problem into many smaller problems that can each be solved individually and then combined to make the overall solution.

    Many sampling-based motion planning algorithms are based on Optimal Rapidly-exploring Random trees (RRT*) \cite{Karaman2011}.
    RRT* iteratively samples the continuous space it plans over building a root search tree from the initial robot location to every other reachable part of the obstacle-free space.
    As RRT* searches, local optimizations are performed on the search tree shortening the length of the paths through the tree.
    As the number of iterations RRT* performs goes to infinity, the resulting solution path from the initial location to the target location converges to optimality \cite{Karaman2011}.
    RRT* and its variants have been shown to solve a large range of path planning problems rapidly.
    However, convergence can be slow \cite{Gammell2014, Nasir2013, Jeong2019}.

    Fast Marching Trees (FMT*) is a sampling-based path planning algorithm that makes use of dynamic programming principles to avoid unnecessary calculations \cite{Janson2015}.
    FMT* builds a rooted search tree similar to RRT*.
    However, instead of iteratively sampling the state space and building the search tree simultaneously, FMT* samples a fixed number of times before starting to build a search tree.
    Once all of the samples have been generated, FMT* builds a search tree with the knowledge of the location of each sample from the beginning.
    Using this knowledge, FMT* is able to avoid many of the calculations that RRT* performs while locally optimizing its search tree.
    This makes FMT* faster than RRT* at generating a solution from a set number of samples.
    However, FMT* is unable to continue refining its solution after that set number of samples.
    RRT*, on the other hand, is purely iterative and will continue to refine its solution path for as long as it is allowed.

    Batch Informed trees (BIT*) combines the iterative nature of RRT* with the efficient graph searching used in FMT* \cite{Gammell2020}.
    BIT* iteratively samples a batch of random samples from the configuration space, then it uses a procedure similar to FMT* to incorporate the new batch of samples into the pre-existing search tree.
    The performance of BIT* varies with the size of each batch of samples.
    When the batch size of BIT* was just one, BIT* is nearly equivalent to RRT*.
    When the batch size is very large, BIT* is nearly equivalent to FMT* except for being able to sample subsequent batches of samples after the first is used.
    This allows the user of the algorithm to tune BIT* to have the performance characteristics desired for a particular application.

    The rest of this work proceeds as follows.
    Section \ref{sec:notation} describes general RRT* and BIT* notation used throughout this work.
    Section \ref{sec:bit} gives a thorough step-by-step explanation of the BIT* algorithm.
    Section \ref{sec:demonstration} provides a demonstration of how BIT* operates, graphically, over three batches of samples.
    Section \ref{sec:simulation} provides averaged converges results using the Open Motion Planning Library (OMPL) and making comparisons to RRT*.

  \section{Notation and Background} \label{sec:notation}
    Section \ref{sec:nomenclature} describes the notation used throughout this work.
    Section \ref{sec:common_procedures} defines some general helper procedures, to be used in the description of BIT*.

    \subsection{Nomenclature} \label{sec:nomenclature}
      RRT* and BIT*-based algorithms iteratively construct a rooted, out-branching tree to find a path through the state space.
      The tree is an acyclic directed graph denoted as $T \triangleq \left(V,E\right)$, where $V$ is the set of nodes or vertices within the tree and $E \subset V \times V$ denotes the set of edges between vertices.
      The root vertex has no parent while all other vertices have exactly one parent.
      Each vertex can have multiple children.
      Each vertex within $V$ corresponds to a state in the $d$-dimensional state space denoted as $X \subset \mathbb{R}^d$.
      The tree is initialized with solely the root node, i.e. $V = \left\{x_{r}\right\}$, $E = \varnothing$.
      Nodes are added to the tree to find a path to the target set, $X_{t} \subset X$.
      The path must avoid the space blocked by obstacles, $X_{obs} \subset X$, staying within the obstacle-free space, $X_{free} \triangleq X \setminus X_{obs}$.

      $\mathbb{R}$ is the set of all real numbers, and $\mathbb{R}_+$ is the set of all real positive numbers.
      The notation $X \xleftarrow{+} \left\{x\right\}$ and $X \xleftarrow{-} \left\{x\right\}$ is used to compactly represent the set updating operations $X \leftarrow X \setminus \left\{x\right\}$ and $X \leftarrow X \cup \left\{x\right\}$ respectively.

    \subsection{Common Procedures} \label{sec:common_procedures}
      In this section, a number of primitive sub-procedures are defined for use while describing BIT*.
      We now define several generic procedures that can be found in \cite{Gammell2020,Swedeen2023} with notation updated to match the sequel.

      \begin{definition}[$cost \leftarrow c\left(x,y\right)$]
        Calculates the length of the path in $X$ that connects $x \in X$ to $y \in X$.
        Note that $c$ returns infinity if the path is blocked by an obstacle.
      \end{definition}

      \begin{definition}[$cost \leftarrow \hat{c}\left(x,y\right)$]
        Calculates a lower bounded heuristic for the cost of the path that goes from $x \in X$ to $y \in X$, i.e., $0 \leq \hat{c}\left(x,y\right) \leq c\left(x,y\right) \leq \infty$.
        Note that $\hat{c}$ typically will not consider obstacles or have to generate the edge explicitly which makes it a fast operation.
        In the work $\hat{c}$ is defined using Euclidean distance as $\hat{c}\left(x,y\right) := \left\|x - y\right\|$.
      \end{definition}

      \begin{definition}[$cost \leftarrow g_T\left(x\right)$]
        Calculates the cost-to-come from the root node to $x \in X$ through the tree, $T$.
        Note that if $x$ is not in the tree then $g_T\left(x\right) = \infty$.
      \end{definition}

      \begin{definition}[$cost \leftarrow \hat{g}\left(x\right)$]
        Calculates a lower bounded heuristic for the cost-to-come of $x \in X$, i.e., $0 \leq \hat{g}\left(x\right) \leq g_T\left(x\right)$.
        In this work $\hat{g}$ is defined using the edge cost heuristic as $\hat{g}\left(x\right) := \hat{c}\left(x_r, x\right)$, where $x_r$ is the initial state of the path planning problem.
      \end{definition}

      \begin{definition}[$cost \leftarrow \hat{h}\left(x\right)$] \label{def:h_hat}
        Calculates a lower bounded heuristic for the cost-to-go of $x \in X$.
        In this work $\hat{h}$ is again defined using edge cost heuristic as $\hat{h}\left(x\right) := \hat{c}\left(x, x_t\right)$, where $x_t \in V$ is the end of the best solution that has been found by the planner so far.
      \end{definition}

      \begin{definition}[$X_{rand} \leftarrow Sample\left(m\right)$]
        Generates a set of $m \in \mathbb{R}_+$ random samples of the obstacle-free state set, $X_{free}$\footnote{Because it is computationally expensive to uniformly sample an arbitrary set, all of $X$ is sampled instead and the sample is discarded and re-sampled if it happens to be in the obstacle set.}.
        The sampling is an independent, identically, and uniformly distributed (i.i.u.d.) sample of $X_{free}$\footnote{It is important that the sampling of $X_{free}$ is i.i.u.d. to guarantee asymptotic optimality \cite{Karaman2011}.}.
      \end{definition}

      \begin{definition}[$X_{near} \leftarrow Near_{\rho}\left(x,X_{search}\right)$]
        Finds all vertices in $X_{search}$ that are within a given edge cost radius\footnote{In this work $\rho$ is held constant, but many sampling-based algorithms vary $\rho$ as the algorithm runs \cite{Karaman2011}.}, $\rho \in \mathbb{R}_+$, of the point $x \in X$, i.e.
        \begin{equation*}
          Near_{\rho}\left(x,X_{search}\right) := \left\{ v \in X_{search} \middle| c\left(x, v\right) \leq \rho \right\}.
        \end{equation*}
      \end{definition}

      \begin{definition}[$X_{sol} \leftarrow Solution\left(x\right)$]
        Finds the path through $T\triangleq\left(V,E\right)$, $X_{sol} \subset V$, that leads from the root node to $x \in V$.
      \end{definition}

      \begin{definition}[$x_{p} \leftarrow Par\left(x_{c}\right)$]
        Returns the parent node of $x_{c} \in V$ in the tree $T\triangleq\left(V,E\right)$, or $\varnothing$ if $x_{c}$ is the root node.
      \end{definition}

      \begin{definition}[$X_{children} \leftarrow Children\left(x_{p}\right)$]
        Returns every node from the set $V$ in $T\triangleq\left(V,E\right)$ that has $x_{p}$ as its parent.
      \end{definition}

      \begin{definition}[$cost \leftarrow BestValue\left(\mathcal{Q}_i\right)$]
        Finds the element in $\mathcal{Q}_i$ with the lowest queue cost and returns the queue cost of that element.
      \end{definition}

      \begin{definition}[$x \leftarrow PopBest\left(\mathcal{Q}_i\right)$]
        Finds the element in $\mathcal{Q}_i$ with the lowest queue cost, removes it from the queue, and returns that element.
      \end{definition}

  \section{Batch Informed Trees} \label{sec:bit}
    \begin{algorithm}[tb]
      \caption{$BIT^*$}
      \label{alg:bit}
      \KwIn{$x_r,X_{goal},X_t$}
      \KwOut{$X_{sol}$}

      $V \leftarrow \left\{x_r\right\}$; \label{alg:bit:start_init}
      $E \leftarrow \varnothing$;
      $T \triangleq \left(V,E\right)$\;
      $\mathcal{Q}_V \leftarrow V$;
      $\mathcal{Q}_E \leftarrow \varnothing$;
      $\mathcal{Q} \triangleq \left(\mathcal{Q}_V,\mathcal{Q}_E\right)$\; \label{alg:bit:init_queues}
      $X_{ncon} \leftarrow X_{goal}$; \label{alg:bit:start_init_flags}
      $X_{new} \leftarrow X_{ncon}$\;
      $V_{exp} \leftarrow \varnothing$;
      $V_{rewire} \leftarrow \varnothing$;
      $V_{sol} \leftarrow V \cap X_t$\;
      \leIf{$V_{sol} = \varnothing$}{$c_{sol} \leftarrow \infty$}{$c_{sol} \leftarrow \min_{v \in V_{sol}} g_T\left(v\right)$}
      $X_{flags} \triangleq \left(X_{new},V_{exp},V_{rewire},V_{sol},c_{sol}\right)$\; \label{alg:bit:end_init}

      \Repeat{STOP}
      { \label{alg:bit:start_loop}
        \If(\tcp*[h]{End of batch}){$\mathcal{Q}_V = \varnothing \land \mathcal{Q}_E = \varnothing$}
        { \label{alg:bit:end_batch}
          \tcp{Prune sub-optimal nodes}
          $\left\{X_{reuse},T,X_{ncon},X_{flags}\right\} \leftarrow Prune\left(T,X_{ncon},X_{flags}\right)$\; \label{alg:bit:call_prune}
          \tcp{Generate new batch of samples}
          $X_{new} \leftarrow Sample\left(m\right)$\; \label{alg:bit:sample}
          \tcp{Add new samples to queues}
          $X_{ncon} \xleftarrow{+} X_{new} \cup X_{reuse}$\; \label{alg:bit:start_add_samples}
          $\mathcal{Q}_V \leftarrow V$\; \label{alg:bit:update_vert_queue}
        }
        \ElseIf{$BestValue\left(\mathcal{Q}_V\right) \leq BestValue\left(\mathcal{Q}_E\right)$}
        { \label{alg:bit:queues_cond}
          \tcp{Process best vertex available}
          $\left\{\mathcal{Q},X_{flags}\right\} \leftarrow ExpVertex\left(T,\mathcal{Q},X_{ncon},X_{flags}\right)$\; \label{alg:bit:exp_vert}
        }
        \Else(\tcp*[h]{Process best edge available})
        {
          $\left\{T,\mathcal{Q},X_{ncon},X_{flags}\right\} \leftarrow ExpEdge\left(T,\mathcal{Q},X_{ncon},X_{flags},X_t\right)$\;
        }
       } \label{alg:bit:end_loop}

       \Return $Solution\left(\argmin_{v_t \in V_{sol}} g_T\left(v_t\right)\right)$;
    \end{algorithm}


    BIT* functions by iteratively generating batches of samples from the state space and incorporating those new samples into the pre-existing search tree.
    To achieve this BIT* defines two sorted queues, the vertex queue and the edge queue.
    At the beginning of each batch, the vertex queue is populated with all of the nodes that are currently in the search tree.
    Vertices are then iteratively removed from the vertex queue and all potential edges that start from that vertex are added to the edge queue.
    Once all potential edges between samples of the state space have been found the search tree is updated to include them if doing so will shorten the length of the resulting path.
    Once both of the queues are empty a new batch is sampled and the process begins again.

    Algorithm \ref{alg:bit} shows the BIT* algorithm.
    The inputs are the root node, $x_r$, a sampling of the target set, $X_{goal} \subset X_t$, and the target set, $X_t \subset X_{free}$.
    Line \ref{alg:bit:start_init} initializes the search tree, $T$, with the root node, $x_r$, in its vertex set, $V$, and no edges in its edge set, $E$.
    Line \ref{alg:bit:init_queues} initializes the vertex queue, $\mathcal{Q}_V$, and the edge queue, $\mathcal{Q}_E$.
    $\mathcal{Q}_V$ is used to keep track of vertices that are under consideration for making potential edges and is organized in terms of the current cost-to-come plus the heuristic cost-to-go of the vertices, i.e.,
    \begin{align}
      g_T\left(v\right) + \hat{h}\left(v\right), v \in \mathcal{Q}_V.
    \end{align}
    $\mathcal{Q}_E$ is used to keep track of edges that are under consideration for addition to $T$.
    $\mathcal{Q}_E$ is organized in terms of the sum of the current cost-to-come of the source vertex of the edge, the heuristic cost of the edge, and the heuristic cost-to-go of the target vertex of the edge, i.e.,
    \begin{align}
      g_T\left(v\right) + \hat{c}\left(v,x\right) + \hat{h}\left(x\right), \left(v,x\right) \in \mathcal{Q}_E.
    \end{align}
    Lines \ref{alg:bit:start_init_flags} through \ref{alg:bit:end_init} initialize a few sets that are needed to keep track of the state of each vertex.
    $X_{ncon} \subset X$ is the set of all samples that are not connected to the search tree.
    Note that $X_{ncon}$ is initialized with the provided samples of the target set, $X_{goal}$.
    $V_{sol} = V \cap X_t$ is the set of vertices in $V$ that are also in the target set $X_t$.
    $X_{new} \subset X_{ncon}$ is the set of samples that are from the most recent batch of samples.
    $V_{nexp} \subset V$ is the set of vertices that have not been considered for expansion.
    $V_{nrewire} \subset V$ is the set of vertices that have not been considered for rewiring.
    $c_i$ is the cost of the current best solution.

    The loop on lines \ref{alg:bit:start_loop} through \ref{alg:bit:end_loop} performs the rest of the planning process and ends when a user-defined stopping condition is met\footnote{Common stopping conditions include achieving a desirable solution cost or expending the extent of the planning time given.}.
    The conditionals on lines \ref{alg:bit:end_batch} and \ref{alg:bit:queues_cond} determine if the batch has ended and whether to process a vertex from $\mathcal{Q}_V$ or an edge from $\mathcal{Q}_E$, respectively.
    Each case is discussed below.

    \subsection{Generating New Batches}
      \begin{algorithm}[tb]
        \caption{$Prune$}
        \label{alg:prune}
        \KwIn{$T \triangleq \left(V,E\right),X_{ncon},X_{flags} \triangleq \left(X_{new},V_{exp},V_{rewire},V_{sol},c_{sol}\right)$}
        \KwOut{$X_{reuse},T,X_{ncon},X_{flags}$}

        $X_{reuse} \leftarrow \varnothing$\;
        $X_{ncon} \xleftarrow{-} \left\{x \in X_{ncon} \middle| \hat{g}\left(x\right) + \hat{h}\left(x\right) \geq c_{sol}\right\}$\; \label{alg:prune:ncon}
        \ForAll{$v \in V$}
        { \label{alg:prune:tree_loop}
          \If{$g_T\left(v\right) + \hat{h}\left(v\right) > c_{sol}$}
          { \label{alg:prune:cur_tree}
            \tcp{Remove $v$ from the search}
            $V \xleftarrow{-} \left\{v\right\}$;
            $E \xleftarrow{-} \left\{\left(Par\left(v\right),v\right)\right\}$\;
            $X_t \xleftarrow{-} \left\{v\right\}$;
            $X_{exp} \xleftarrow{-} \left\{v\right\}$;
            $X_{rewire} \xleftarrow{-} \left\{v\right\}$\;
            \tcp{If $v$ can possibility help}
            \lIf{$\hat{g}\left(v\right) + \hat{h}\left(v\right) < c_{sol}$}
            { \label{alg:prune:final_con}
              $X_{reuse} \xleftarrow{+} \left\{v\right\}$\!\!\ \label{alg:prune:tree_loop_end}
            }
          }
        }

        \Return $\left\{X_{reuse},T,X_{ncon},X_{flags}\right\}$;
      \end{algorithm}

      When $\mathcal{Q}_V$ and $\mathcal{Q}_E$ are both empty it signifies the end of the batch, see line \ref{alg:bit:end_batch} of Algorithm \ref{alg:bit}.
      When that happens, all vertices that cannot contribute to the optimal solution are removed from the search tree, $T$, by calling the $Prune$ procedure on line \ref{alg:bit:call_prune}.

      The $Prune$ procedure is given in Algorithm \ref{alg:prune}.
      Line \ref{alg:prune:ncon} of Algorithm \ref{alg:prune} removes all unconnected samples with heuristic cost-to-come plus heuristic cost-to-go value greater than the current best solution.
      Note that this can be thought of as removing all nodes that fall outside of the \enquote{informed set} that Informed RRT* (\mbox{I-RRT*}) defines\cite{Gammell2014} and as such provably cannot contribute to the optimal solution.
      The loop on lines \ref{alg:prune:tree_loop} though \ref{alg:prune:tree_loop_end} removes any vertices in the search tree that cannot contribute to the optimal solution.
      Line \ref{alg:prune:cur_tree} checks if each vertex in the tree has the potential to contribute to the optimal solution given its current connection to the search tree.
      If this check fails, the vertex is removed from the search tree.
      Line \ref{alg:prune:final_con} checks if the vertex has the potential to contribute to the optimal solution given the ideal cost-to-come.
      This is effectively the same condition that is used on line \ref{alg:prune:ncon}.
      If there is a chance of the sample contributing to the optimal solution, the vertex is added to $X_{reuse}$ to be reused as an unconnected sample in the next batch.
      $X_{reuse}$ is added to the algorithm to maintain uniform sample density in the \enquote{informed set}.

      After $Prune$ is finished, line \ref{alg:bit:sample} of Algorithm \ref{alg:bit} generates $m$ \emph{new} samples of the obstacle-free state space, $X_{free}$.
      Line \ref{alg:bit:start_add_samples} adds the new samples and reused vertices to $X_{ncon}$.
      Line \ref{alg:bit:update_vert_queue} adds all vertices in the search tree to the vertex queue.
      This insures all vertices in the search tree will be considered when looking for ways to connect the new samples to the search tree.

    \subsection{Expanding Vertices}
      \begin{algorithm}[tb]
        \caption{$ExpVertex$}
        \label{alg:exp_v}
        \KwIn{$T \triangleq \left(V,E\right),\mathcal{Q} \triangleq \left(\mathcal{Q}_V,\mathcal{Q}_E\right),X_{ncon},X_{flags} \triangleq \left(X_{new},V_{exp},V_{rewire},V_{sol},c_{sol}\right)$}
        \KwOut{$\mathcal{Q},X_{flags}$}

        $v_b \leftarrow PopBest\left(\mathcal{Q}_V\right)$\; \label{alg:exp_v:pop}
        \tcp{Make edges to unconnected samples}
        \If{$v_b \not\in V_{exp}$}
        { \label{alg:exp_v:first_time_cond}
          $V_{exp} \xleftarrow{+} \left\{v_b\right\}$\;
          $X_{near} \leftarrow Near_{\rho}\left(v_b,X_{ncon}\right)$\; \label{alg:exp_v:all_uncon_search}
        }
        \Else(\tcp*[h]{$v_b$ has been expanded before})
        {
          $X_{near} \leftarrow Near_{\rho}\left(v_b,X_{new} \cap X_{ncon}\right)$\; \label{alg:exp_v:new_uncon_search}
        }
        $\mathcal{Q}_E \xleftarrow{+} \left\{\left(v_b,x\right), x \in X_{near} \middle| \hat{g}\left(v_b\right) + \hat{c}\left(v_b,x\right) + \hat{h}\left(x\right) < c_{sol}\right\}$\; \label{alg:exp_v:exp_edges}
        \tcp{If $v_b$ has not been rewired before}
        \If{$v_b \not\in V_{rewire} \land c_{sol} < \infty$}
        { \label{alg:exp_v:rewire_cond}
          \tcp{Make edges to connected nodes}
          $V_{rewire} \xleftarrow{+} \left\{v_b\right\}$\;
          $V_{near} \leftarrow Near_{\rho}\left(v_b,V\right)$\; \label{alg:exp_v:in_tree_search}
          $\begin{aligned}
            \mathcal{Q}_E \xleftarrow{+}
              &\left\{
                \left(v_b,w\right), w \in V_{near} \middle| \right.
                   \left(v_b,w\right) \not\in E, \\
                  &\hat{g}\left(v_b\right) + \hat{c}\left(v_b,w\right) < g_T\left(w\right), \\
                  &\left. \hat{g}\left(v_b\right) + \hat{c}\left(v_b,w\right) + \hat{h}\left(w\right) < c_{sol}
              \right\};
          \end{aligned}$\ \label{alg:exp_v:rewire_edges}
        }

        \Return $\left\{\mathcal{Q},V_{flags}\right\}$;
      \end{algorithm}

      Lines \ref{alg:bit:queues_cond} and \ref{alg:bit:exp_vert} of Algorithm \ref{alg:bit} find potential edges to add to $\mathcal{Q}_E$ from the vertices in $\mathcal{Q}_V$.
      The condition on line \ref{alg:bit:queues_cond} evaluates to true until it is impossible for the best vertex in $\mathcal{Q}_V$ to produce an edge of lower heuristic cost then the best edge in $\mathcal{Q}_E$.
      This can be seen by noting that
      \begin{align}
        \forall v \in \mathcal{Q}_V, \forall x \in X, g_T\left(v\right) + \hat{h}\left(v\right) \leq g_T\left(v\right) + \hat{c}\left(v,x\right) + \hat{h}\left(x\right)
      \end{align}
      as $\hat{h}\left(v\right)$ is an under estimate of the true cost-to-go of vertex $v$.
      Thus, the vertex queue cost, $g_T\left(v\right) + \hat{h}\left(v\right)$, is a lower bound on the edge queue cost, $g_T\left(v\right) + \hat{c}\left(v,x\right) + \hat{h}\left(x\right)$ of any edge that can be made from that vertex.

      $ExpVertex$ removes the lowest cost vertex in $\mathcal{Q}_V$ and adds edges to $\mathcal{Q}_E$ for every neighbor that might be part of the optimal solution.
      The $ExpVertex$ procedure is given in Algorithm \ref{alg:exp_v}.
      Line \ref{alg:exp_v:pop} of Algorithm \ref{alg:exp_v} pops the lowest cost vertex in $\mathcal{Q}_V$.
      Lines \ref{alg:exp_v:first_time_cond} through \ref{alg:exp_v:exp_edges} adds edges to $\mathcal{Q}_E$ that start from $v_b$ and go to samples that are not connected to the tree.
      The condition on line \ref{alg:exp_v:first_time_cond} checks if this is the first time $v_b$ has been considered for expansion.
      In that case, all unconnected samples are considered for connection to $v_b$, see line \ref{alg:exp_v:all_uncon_search}.
      If it is not the first time $v_b$ has been considered for expansion, only the samples that are new this batch are considered for connection, see line \ref{alg:exp_v:new_uncon_search}.
      This prevents redundant calculations as the samples that are not new this batch have already been considered for connection to $v_b$.
      Once $X_{near}$ has been found, line \ref{alg:exp_v:exp_edges} adds all edges in the set $\left\{\left(v_b,x\right), x \in X_{near}\right\}$ that have potential to improve the current solution to $\mathcal{Q}_E$.

      Lines \ref{alg:exp_v:rewire_cond} through \ref{alg:exp_v:rewire_edges} handle the case when $v_b$ might be a better parent for its neighbors then their current parent, i.e., rewiring.
      Note that if BIT* has not found a solution the condition on line \ref{alg:exp_v:rewire_cond} will always evaluate to false.
      This is done to reduce the time it takes to find an initial solution to the problem by skipping any potential tree rewirings.
      Once the first solution is found, all rewirings that were skipped previously are considered.
      The condition on line \ref{alg:exp_v:rewire_cond} also evaluates to false if this is not the first time $v_b$ has been considered for rewiring.
      This prevents redundant calculations as the potential to perform rewirings around $v_b$ has already been considered.
      Line \ref{alg:exp_v:in_tree_search} finds all vertices in the search tree that are near $v_b$.
      Line \ref{alg:exp_v:rewire_edges} adds all edges from $v_b$ to $V_{near}$ that are not already part of the tree, have the potential to improve the cost of the neighbor, and have the potential to improve the current solution.

      Note that everything done in $ExpVertex$ is done completely with heuristic values and without obstacle checking.
      This keeps the procedure computationally lightweight and fast.

    \subsection{Evaluating Possible Edges}
      \begin{algorithm}[tbh]
        \caption{$ExpEdge$}
        \label{alg:exp_e}
        \KwIn{$T \triangleq \left(V,E\right),\mathcal{Q} \triangleq \left(\mathcal{Q}_V,\mathcal{Q}_E\right),X_{ncon},X_{flags} \triangleq \left(X_{new},V_{exp},V_{rewire},V_{sol},c_{sol}\right),X_t$}
        \KwOut{$T,\mathcal{Q},X_{ncon},X_{flags}$}

        $\left(v_b,x_b\right) \leftarrow PopBest\left(\mathcal{Q}_E\right)$\; \label{alg:exp_e:pop}
        \If{$g_T\left(v_b\right) + \hat{c}\left(v_b,x_b\right) + \hat{h}\left(x_b\right) \geq c_{sol}$}
        { \label{alg:exp_e:has_any_hope}
          \tcp{The best edge in the edge queue cannot improve the solution}
          $\mathcal{Q}_E \leftarrow \varnothing$;
          $\mathcal{Q}_V \leftarrow \varnothing$\; \label{alg:exp_e:clear}
          \Return $\left\{T,\mathcal{Q},X_{ncon},X_{flags}\right\}$;
        }
        \If(\tcp*[h]{$x_b$ is unconnected}){$x_b \in X_{ncon}$}
        { \label{alg:exp_e:rewire_cond}
          \If{$g_T\left(v_b\right) + c\left(v_b,x_b\right) + \hat{h}\left(x_b\right) < c_{sol}$}
          { \label{alg:exp_e:add_vert_cond}
            \tcp{Add $x_b$ to the tree}
            $X_{ncon} \xleftarrow{-} \left\{x_b\right\}$;
            $V \xleftarrow{+} \left\{x_b\right\}$;
            $E \xleftarrow{+} \left\{\left(v_b,x_b\right)\right\}$\;
            $\mathcal{Q}_V \xleftarrow{+} \left\{x_b\right\}$\;
            \If{$x_b \in X_t$}
            { \label{alg:exp_e:check_targ}
              $V_{sol} \xleftarrow{+} \left\{x_b\right\}$; \label{alg:exp_e:add_targ}
              $c_{sol} \leftarrow \min_{v_{sol} \in V_{sol}} g_T\left(v_{sol}\right)$\;
            }
          }
        }
        \Else(\tcp*[h]{$x_b$ is connected})
        {
          \If{$g_T\left(v_b\right) + \hat{c}\left(v_b,x_b\right) < g_T\left(x_b\right)$}
          { \label{alg:exp_e:hur_imp_neih}
            \If{$g_T\left(v_b\right) + c\left(v_b,x_b\right) + \hat{h}\left(x_b\right) < c_{sol}$}
            { \label{alg:exp_e:help_sol_cond}
              \If{$g_T\left(v_b\right) + c\left(v_b,x_b\right) < g_T\left(x_b\right)$}
              { \label{alg:exp_e:imp_neih}
                $E \xleftarrow{-} \left\{\left(Par\left(x_b\right),x_b\right)\right\}$;
                $E \xleftarrow{+} \left\{\left(v_b,x_b\right)\right\}$\;
                $c_{sol} \leftarrow \min_{v_{sol} \in V_{sol}} g_T\left(v_{sol}\right)$\;
              }
            }
          }
        }

        \Return $\left\{T,\mathcal{Q},X_{ncon},X_{flags}\right\}$;
      \end{algorithm}

      In the case that the best heuristic cost edge possible has been generated from $\mathcal{Q}_V$, the condition on line \ref{alg:bit:queues_cond} of Algorithm \ref{alg:bit} evaluates to false and $ExpEdge$ is called.
      $ExpEdge$ removes the most promising edge from $\mathcal{Q}_E$ and considers it for addition to the search tree.

      The $ExpEdge$ procedure is given in Algorithm \ref{alg:exp_e}.
      Line \ref{alg:exp_e:pop} removes the lowest queue cost edge from $\mathcal{Q}_E$.
      Line \ref{alg:exp_e:has_any_hope} checks if there is a chance that the edge under consideration will improve the current solution.
      Note that this condition is true only if the most promising edge in $\mathcal{Q}_E$, and by extension all of the edges in $\mathcal{Q}_E$, cannot contribute to the optimal solution.
      For this reason $\mathcal{Q}_E$ and $\mathcal{Q}_V$ are cleared on line \ref{alg:exp_e:clear}.

      Line \ref{alg:exp_e:rewire_cond} checks if $x_b$ is already in the tree.
      If $x_b$ is not part of the search tree, line \ref{alg:exp_e:add_vert_cond} checks if connecting $x_b$ to the tree through $v_b$ can improve the current solution.
      If it can, $x_b$ is added to the search tree with $v_b$ as its parent.
      Lines \ref{alg:exp_e:check_targ} and \ref{alg:exp_e:add_targ} check if $x_b$ is in the target set and adds $x_b$ to $V_{sol}$ if so.
      If $x_b$ is already part of the search tree at line \ref{alg:exp_e:rewire_cond}, extra checks are performed before adding the edge under consideration to the tree.
      Line \ref{alg:exp_e:hur_imp_neih} checks if connecting $x_b$ through $v_b$ can improve the cost of $x_b$.
      Line \ref{alg:exp_e:help_sol_cond} checks that the edge under consideration can improve the current solution.
      Line \ref{alg:exp_e:imp_neih} checks that connecting $x_b$ through $v_b$ will improve the cost of $x_b$.
      If all checks pass, $x_b$ is rewired to have $v_b$ as its parent and the current solution cost is updated if needed.

  \section{Demonstration} \label{sec:demonstration}
    \input{tikz_pics/bit_demo.tex}

    Figures \ref{fig:bit_demo:batch0} through \ref{fig:bit_demo:batch3} graphically show how BIT* operates over three batches of samples.
    In this example, each batch consists of five samples, i.e., $m = 5$.

    Figure \ref{fig:bit_demo:batch0} shows what we call batch 0.
    This is the part of the planning process where the only node in $\mathcal{Q}_V$ is the root node and no samples have been made from the state space.
    This part of the algorithm checks for the trivial case where it is possible to directly connect the root node to the target set.

    Figure \ref{fig:bit_demo:batch1} shows the process of generating the first batch of samples and starting to build a search tree.
    Note that when batch 1 completes a path to the target set has not been found.
    This is a common occurrence in BIT* where the search tree is unable to grow much until the sample density in $X_{free}$ grows for a few batches.

    Figure \ref{fig:bit_demo:batch2} shows how the second batch of samples is incorporated into the search tree.
    By the end of the batch a path to the target set has been found and the \enquote{informed ellipse} is defined.
    This enables pruning to be performed before batch 3 begins.

    Figure \ref{fig:bit_demo:batch3} shows how a new batch of samples is generated within the informed ellipse and used to refine the current solution.
    Note that as the solution length reduces in the third batch, the size of the informed ellipse also shrinks.
    This leads to more nodes being pruned and the search process becoming more focused on the area that can improve the current solution.

  \section{Simulation} \label{sec:simulation}
    Simulation results are now presented to demonstrate the effectiveness of the BIT* algorithm.
    Comparisons are made between BIT* and RRT* planning with straight-lines.

    \subsection{Simulation Details}
      \begin{figure*}[tbh]
        \centering
        \includegraphics[width=0.8\linewidth]{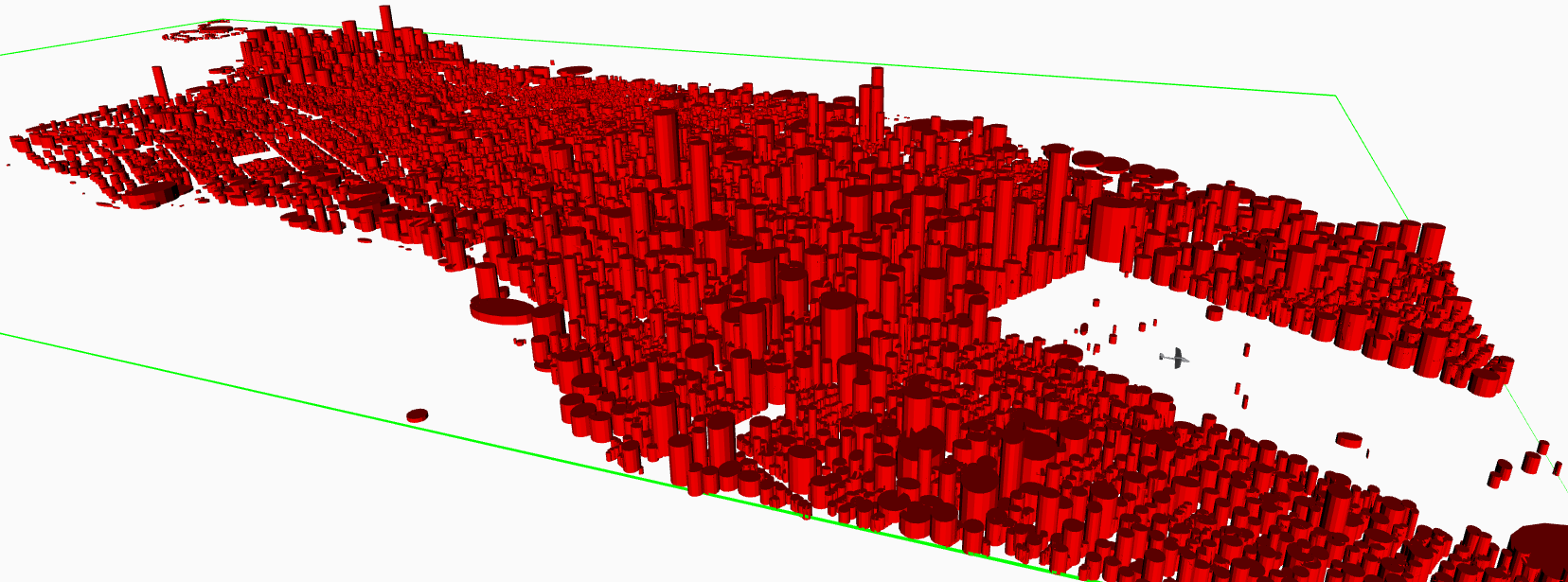}
        \caption{The UAV simulation with Manhattan's buildings shown in red.}
        \label{fig:uav_sim}
      \end{figure*}
      \input{tikz_pics/worlds.tex}

      To test the capabilities of BIT*, it is used to plan paths for a simulated UAV through an urban environment.
      The UAV simulation is shown in Figure \ref{fig:uav_sim}.
      The buildings in the UAV simulation are modeled off of the real buildings in Manhattan, New York.
      The placement and height of the buildings are from New York's Open Data project \cite{NycOpenData2016}.
      The initial position of the UAV is located in Central park.
      Maintaining an altitude of $10$ meters, the UAV plans a path through the buildings to the goal location on Governors Island.

      While planning a path, obstacles are represented with an occupancy grid with each pixel corresponding to ten square centimeters.
      Every building that is taller than $10 m$ is considered an obstacle in the occupancy grid.
      Figure \ref{fig:all_worlds} shows the resulting occupancy grid with black representing obstacles.
      The occupancy grid covers a $10.7 km \times 5.65 km$ area of Manhattan.
      The initial location of the UAV in the coordinate frame used for this problem is $x_r = \begin{bmatrix} 0 & 0 \end{bmatrix} km$, with the UAV orientation defined in the direction of the target set.
      The target set, $X_{t}$, is a circle of radius $0.001m$ centered at $\begin{bmatrix} -9 & -3.8 \end{bmatrix} km$.
      The samples of the target set that are provided to BIT*, $X_{goal}$, is the singleton set of the center of $X_t$.
      The neighborhood search radius, $\rho$, is $500 m$.
      Paths are generated and checked for obstacles four times per every meter of path length.
      When using BIT*, the batch size, $m$, is set to $1500$ samples.

      When using RRT* there are three additional parameters, $\eta$, $\alpha$, and $b_t$, that are described in \cite{Swedeen2023} but only given values here for brevity.
      The steering constant, $\eta$, is $500 m$.
      The max number of neighbors to search, $\alpha$, is $100$ neighbors.
      The check target period, $b_t$, is $1$ out of every $50$ samples.

      Results are gathered using the Open Motion Planning Library (OMPL) \cite{Moll2015}.
      As the sampling is random, each simulation consists of over 100 individual simulations with the average results being presented.
      The results were gathered on an AMD Ryzen\texttrademark{} Threadripper\texttrademark{} 3970X processor.
      Convergence plots were made by fitting a 15\textsuperscript{th}-order polynomial using a least-squares fitting algorithm as described in \cite{Venables2002}.

      A least-squares approach is used as the sampling times for path length are not uniform across all simulations and not all simulations find the initial path at the same time.
      Note that while the path length for any one run will be monotonically decreasing with time, the least squares fitted plot does not always have the same monotonic property.
      The reason is that a particular run may not find a solution until well after other runs and the initial solution it finds may be much larger than the current solution of the other runs, effectively causing the average to increase at the time the run first produces path length data.

      Simulation code can be found in our open-source repository \url{https://gitlab.com/utahstate/robotics/fillet-rrt-star} \TODO{Update it with BIT* code}.

    \subsection{Results}
      \input{tikz_pics/convergence_plots.tex}

      The convergence results from benchmarking RRT* and BIT* in the Manhattan environment are shown in Figure \ref{fig:transients}.
      RRT* and BIT* are shown in blue and red respectively.
      Clearly BIT* outperforms RRT* in terms of converging to a near optimal-value rapidly.
      BIT* initially finds a solution much shorter than RRT* and proceeds to converge to near optimality by the time 30 seconds have passed.
      This comes from BIT*'s use of cost-to-come and cost-to-go heuristics to focus their search efforts in directions that are most likely to improve the solution cost.
      RRT* on the other hand starts with a long initial solution.
      Despite converging for five minutes, the solution that RRT* produces is still substantially longer than that of BIT*.


  \bibliographystyle{IEEEtran}
  \bibliography{refs}
\end{document}

%% file: tikz_pics/worlds.tex
\tikzsetnextfilename{worlds}
\begin{figure*}[tbh]
  \newcommand{\W}{\linewidth}
  \begin{subfigure}[t]{\W}
    \centering
    \resizebox{\linewidth}{!}{
      \begin{tikzpicture}
        \begin{axis}[
          enlargelimits=false,
          axis lines=none,
          xmin=-9700,ymin=-4650,xmax=1000,ymax=1000,
          scale only axis,
          axis equal image,
          axis equal=true,
          ]
          \addplot graphics [xmin=-9700,ymin=-4650,xmax=1000,ymax=1000] {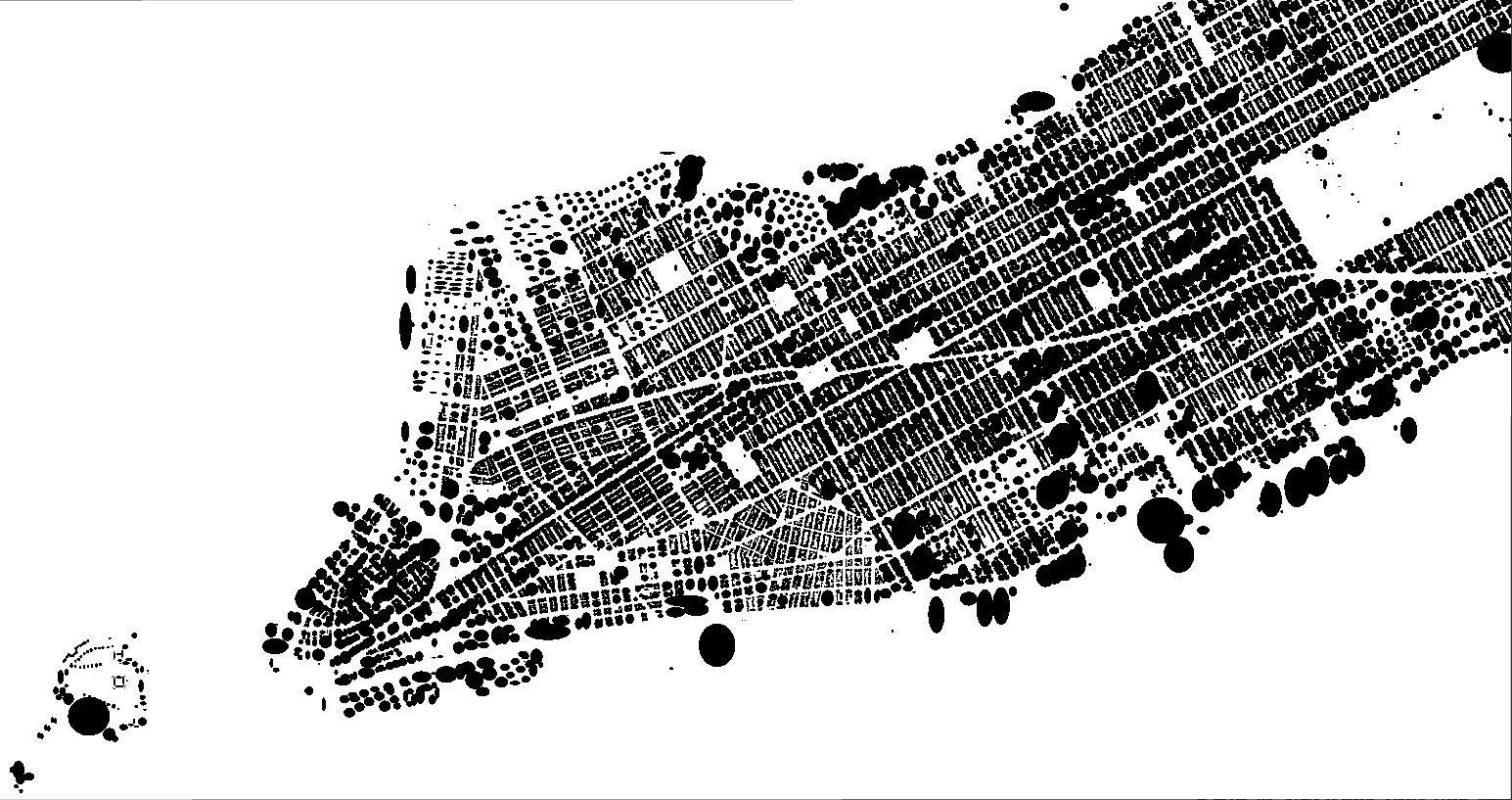};

          \addplot[color=red]   table [x=x,y=y, col sep=comma] {pics/manhattan/straight.csv};

          \node[label={right:{\small$x_r$}},inner sep=0pt] () at (0,0) {};
          \node[label={above:{\small$X_t$}},inner sep=0pt] () at (-9000,-3700) {};
          \draw[thick] (-9700,-4650) -- (-9700,1000) -- (1000,1000) -- (1000,-4650) -- cycle;
        \end{axis}
      \end{tikzpicture}
    }
  \end{subfigure}
  \caption{The resulting paths from running BIT* in the Manhattan world.
           The Path from BIT* is shown in red.}
  \label{fig:all_worlds}
\end{figure*}

%% file: tikz_pics/convergence_plots.tex
\tikzsetnextfilename{convergence_plots}
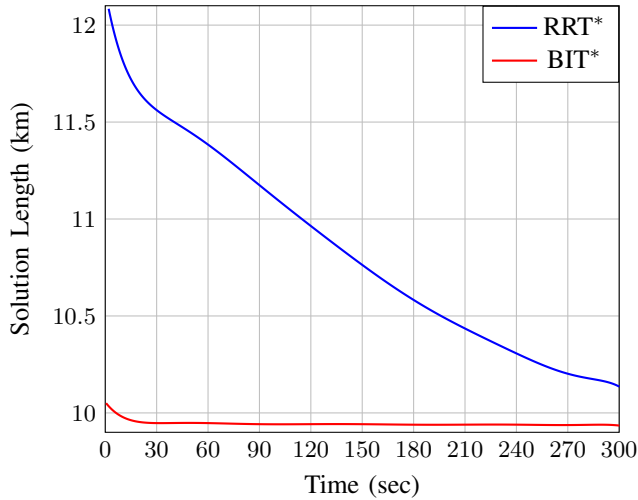
\begin{figure}[tbh]
  \centering
    \begin{tikzpicture}
      \begin{groupplot}[
        group style={
          group size=1 by 1,
          vertical sep=5pt,
          horizontal sep=2pt,},
        ymin=9900,
        ymax=12100,
        xmin=0,
        xmax=300,
        width=0.95\linewidth,
        ticklabel style={font=\small},
        grid=both,
        xlabel={Time (sec)},
        ytick style={opacity=0},
        xtick style={opacity=0},
        ylabel near ticks,
        xlabel near ticks,
        legend style={nodes={scale=1, transform shape},at={(1,1)},anchor=north east},
        change y base=true,
        y SI prefix=kilo,
        xtick distance=30,
        ]

        \nextgroupplot[
          title={Convergence Trends},
          ylabel={Solution Length (km)},
          ]

        \addplot[color=blue, thick] table [x=x,y=y, col sep=comma] {sim_data/holonomic_rrt_star.csv};
        \addplot[color=red,  thick] table [x=x,y=y, col sep=comma] {sim_data/holonomic_bit.csv};

        \legend{RRT$^*$, BIT$^*$}

      \end{groupplot}
    \end{tikzpicture}
  \caption{The convergence plots of the Manhattan world over 5 minutes.
           RRT* and BIT* planning with straight-line motion primitives are shown in blue and red respectively.}
  \label{fig:transients}
\end{figure}